\documentclass[journal]{IEEEtran}

\usepackage{times}
\usepackage{epsfig}
\usepackage{graphicx}
\usepackage{amsmath}
\usepackage{amssymb}
\usepackage{booktabs}

\usepackage{xcolor}
\usepackage{enumitem}
\usepackage{lipsum}
\usepackage{multirow}

\usepackage{tcolorbox}
\usepackage{colortbl}
\definecolor{lavender}{rgb}{0.9, 0.9, 0.98}
\definecolor{magenta}{rgb}{0.96, 0.0, 0.8}

\usepackage{pifont}
\newcommand{\cmark}{\ding{51}}%
\newcommand{\xmark}{\ding{55}}%
\newcommand{\ie}{{\it i.e.}}
\newcommand{\eg}{{\it e.g.}}

\usepackage[breaklinks=true,bookmarks=false]{hyperref}

\begin{document}

\title{The Second-place Solution for ECCV 2022 Multiple People Tracking in Group Dance Challenge}

\author{
Fan Yang, Shigeyuki Odashima, Shoichi Masui, Shan Jiang\\
Fujitsu Research, Japan\\
{\tt\small contact: hongheyangfan@gmail.com; fan.yang@fujitsu.com}
}
\maketitle
\thispagestyle{empty}

\begin{abstract}
   This is our $2^{nd}$-place solution for the ECCV 2022 Multiple People Tracking in Group Dance Challenge. Our method mainly includes two steps: online short-term tracking using our Cascaded Buffer-IoU (C-BIoU) Tracker, and, offline long-term tracking using appearance feature and hierarchical clustering.
   Our C-BIoU tracker adds buffers to expand the matching space of detections and tracks, which mitigates the effect of irregular motions in two aspects: one is to directly match identical but non-overlapping detections and tracks in adjacent frames, and the other is to compensate for the motion estimation bias in the matching space. In addition, to reduce the risk of overexpansion of the matching space, cascaded matching is employed: first matching alive tracks and detections with a small buffer, and then matching unmatched tracks and detections with a large buffer. After using our C-BIoU for online tracking, we applied the offline refinement introduced by ReMOTS~\cite{yang2020remots}.
\end{abstract}

\vspace{-3.5mm}
\section{Introduction}

\begin{figure}[th!]
   \centering
   \includegraphics[width=\linewidth]{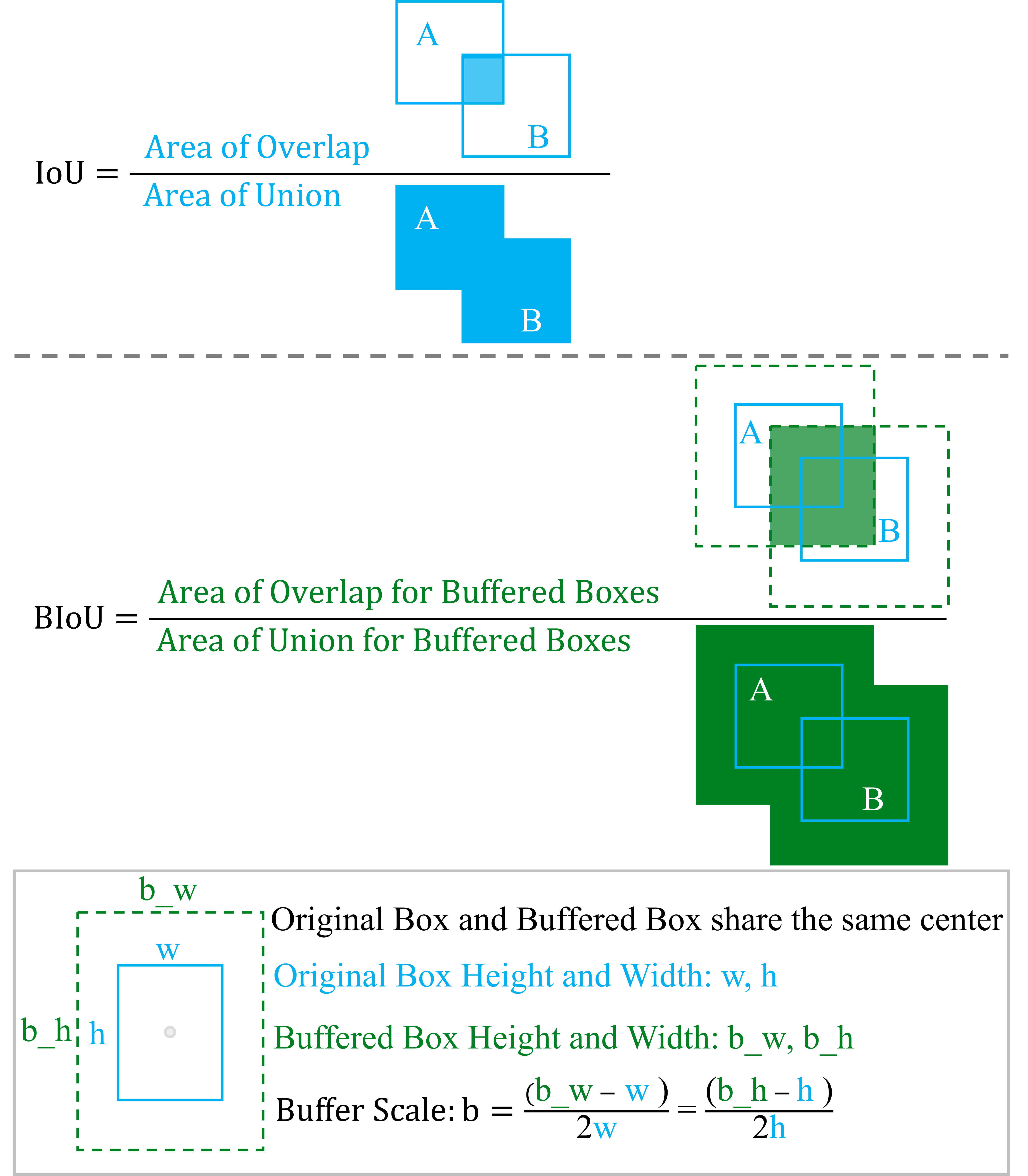}
   \caption{\textbf{Illustration of how Buffered IoU (BIoU) is calculated}. Our BIoU adds a buffer that is proportional to the original bounding box. It does not change the location center, scale ratio, and shape of the original bounding boxes but expands the original matching space.}
   \label{fig:BIoU}
\end{figure}

\begin{figure*}[h!]
   \centering
   \includegraphics[width=\linewidth]{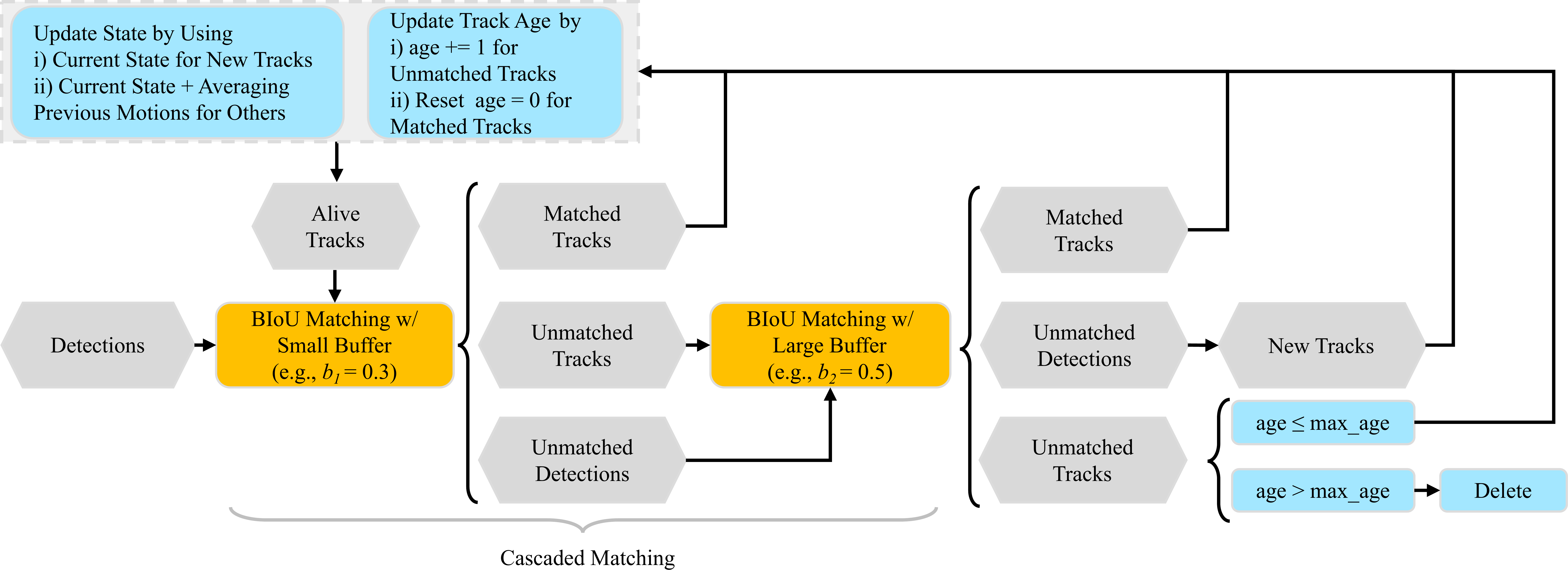}
   \caption{\textbf{Framework of Our C-BIoU Tracker.} }
   \label{fig:C-BIoU_framework}
\end{figure*}

Although MOT studies have been greatly developed~\cite{SORT,DeepSORT,bytetrack, Li_2022_CVPR},  a new challenge has recently attracted attention: unlike conventional MOT tasks that contain objects with distinct appearances and regular motions, MOT tasks that cover animals, group dancers, and sports players, may have indistinguishable appearances and irregular motions, which could cause existing MOT methods to fail. In particular, several MOT methods~\cite{SORT, DeepSORT, bytetrack, cao2022observation} that perform well on MOT17~\cite{MOT16}, may experience a significant performance drop on the DanceTrack~\cite{sun2022dancetrack}.

We presume that tracking failures are caused by two reasons: 
(i) The detections and tracks of identical objects do not overlap between adjacent frames (\eg, due to the fast movement) and thus the tracking fails; (ii) After track initialization, unmatched tracks (\eg, occluded objects) continue to update their geometric features for multiple frames, however, if their motion estimations are inaccurate (\eg, due to a sudden acceleration or turning), they miss the matching opportunity when corresponding detections are available in subsequent frames. When the appearance of objects can be distinguished, appearance features could be employed to alleviate issues (i) and (ii), by matching cross-frame detections based on their appearance similarities. Nonetheless, when irregular motions are accompanied by indistinguishable appearances, most existing MOT solutions may not be able to perform dependable tracking, so a new solution is desirable.

\begin{figure}[t!]
   \centering
   \includegraphics[width=0.6\linewidth]{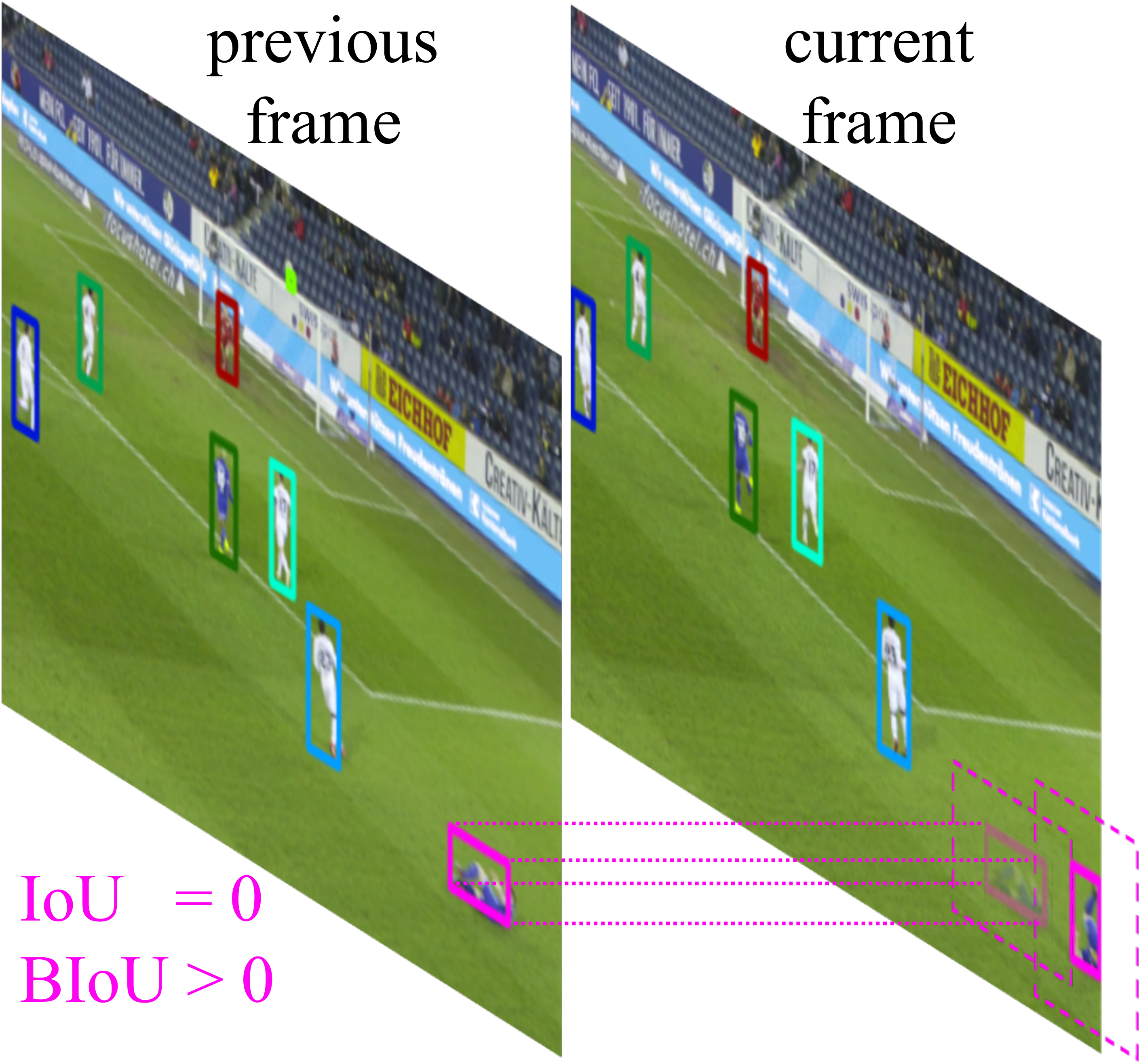}
   \caption{\textbf{An illustration of BIoU forms better cross-frame geometric consistency than IoU}. The bounding box of an identical object shares the same color. The \textcolor{magenta}{magenta object} has no overlapping detections between adjacent frames. Whether this is caused by the fast movement or incorrect motion estimation, our BIoU expands the matching space to reduce the miss matching.}
   \label{fig:BIoU_demo}
\end{figure}

In this study, we propose a Cascaded-Buffered Intersection over Union (C-BIoU) tracker~\cite{CBIOU_2023_WACV} to track multiple objects that have irregular motions and indistinguishable appearances.


\section{Method}

\subsection{Detection of C-BIoU Tracker}

Our C-BIoU tracker (Fig.~\ref{fig:C-BIoU_framework}) follows the tracking-by-detection paradigm---the object detector and MOT framework are separately designed. Given a video, we applied YOLOX~\cite{ge2021yolox} trained by OC-SORT~\cite{cao2022observation} to generate bounding boxes at each frame. We tried input sizes of $[800, 1440]$ and $[960, 1728]$.
Our C-BIoU tracker then takes detected bounding boxes as inputs to produce tracking results. Such a pipeline provides great flexibility to apply our C-BIoU tracker on arbitrary detections.

\subsubsection{Buffered IoU}

The Buffered IoU (BIoU) is our main contribution in this work. As shown in Fig.~\ref{fig:BIoU}, the BIoU simply adds buffers that are proportional to the original detections and tracks for calculating the IoU. Our BIoU retains the same location centers, scale ratios, and shapes of the original detections and tracks, but it expands the matching space to measure the geometric consistency. Let $\boldsymbol{o} = (x, y, w, h)$ denote an original detection and $(x,y,w,h)$ be the top-left coordinate, width, and height of the detection, respectively. Suppose that the buffer scale is $b$, we have the buffered detection as $\boldsymbol{o}_{b} = ( x-bw,y-bh, w+2bw, h+2bh )$. To approach our cascaded matching, we apply grid research~\cite{bergstra2012random} to find the best combination of two buffer scales $b_{1}$ and $b_{2}$ on the training set, and then apply them to the validation set and test set. Since we have $b_{1}<b_{2}$, when we search for the combination of $b_{1}$ and $b_{2}$ within a certain range, the number of combinations is limited. Considering that the speed of our C-BIoU is fast, the grid search takes an acceptable time.

\begin{table*}[t]
   \caption{\textbf{Result details of our method in the ECCV 2022 Group Dance Challenge.} Using non-maximum suppression (NMS), we merged detections generated by two input scales. The data rendered in \textbf{Bold} indicate the best results.}
   \centering
   \resizebox{\linewidth}{!}{
   \begin{tabular}{ l | l| ccccc}
   \toprule
   Tracker & Detector &HOTA$\uparrow$ & DetA$\uparrow$ & AssA$\uparrow$ & MOTA$\uparrow$ & IDF1$\uparrow$\\
   \midrule 
   C-BIoU (Online) & YOLOX-X $[800, 1440]$&  60.6 & \textbf{81.3} & 45.4 & 91.6 & 61.6\\
   C-BIoU (Online) & Merged YOLOX-X $[800, 1440]$ and $[960, 1728]$ &  61.7  &  \textbf{81.3}  & 47.0 & 92.2 & 65.3\\
   C-BIoU (Online) + ReMOTS (Offline Refinement) &Merged YOLOX-X $[800, 1440]$ and $[960, 1728]$  &  \textbf{66.6} & \textbf{81.3} & \textbf{54.7} & \textbf{92.3} &	\textbf{71.4} \\
   \bottomrule
   \end{tabular}
   }
   \label{table:dancetrack_challenge}
 \end{table*}

\subsubsection{Simple Motion Estimation}

Unlike most MOT methods~\cite{SORT,DeepSORT,bytetrack} that apply the Kalman filter ~\cite{kalman1960new} for state estimation, we simply average motions of recent frames to quickly respond to unpredictable motion changes. 
At frame $t$, suppose that a track has matched detections for more than $n$ frames, after $\Delta$ unmatched frames, its track state $\boldsymbol{s}$ can be represented as $\boldsymbol{s}^{t+\Delta} = o^{t} + \frac{\Delta}{n} \sum^{t}_{i=t-n+1} (\boldsymbol{o}^{i}- \boldsymbol{o}^{i-1})$. The matched detections between frame $t-n$ to $t$ are used to calculate motions and the average motion is applied to update the track state. We set $2\leq n \leq 5$ by default in our experiments. The IoU score of buffered $\boldsymbol{s}^{t+\Delta}_{b}$ and $\boldsymbol{o}^{i+\Delta}_{b}$ is used for data association at the frame $t+\Delta$. Due to the simplicity of our approach, the overall tracking speed is increased for our C-BIoU tracker.

\subsubsection{Track Management}

In an MOT framework, the function of track management is to decide how and when to initialize, update and terminate a track. We design our track management based on the mainstream solution introduced by SORT~\cite{SORT}. It initializes tracks from unmatched detections, applies the alive tracks to match new detections, and terminates a track when it has not been matched for a given amount of frames (\ie, $max\_age$). Two BIoUs, which respectively equip small and large buffers, are grouped into a cascaded matching. First, we match alive tracks and detections with the BIoU that has a small buffer (\ie, $b_1$). Then, we continue to match unmatched tracks and detections with the BIoU that has a large buffer (\ie, $b_2$). For the motion estimation, we simply average the speeds of recent frames to quickly respond to unpredictable motion changes.

\subsection{Offline Refinement}

Due to the occlusion, some long-term tracklets could be broken down by only referring to geometry features. To recover long-term tracklets, we employed an offline tracking with appearance features. We utilized Strong ReID~\cite{luo2019bag} to obtain appearance feature. To initialize the re-id model, we utilized the DanceTrack training set. To include the tracking labels in our re-id training, we took a self-supervised learning method introduced by ReMOTS~\cite{yang2020remots}. Referring to tracking ID, in each video, we construct triplets and only apply triplet loss to them. 

After the training, we generated appearance features for each short-term tracklets obtained from the previous step. Within a video, we formed a distance matrix $\mathcal{D}$ between short-term tracklets as
\begin{equation}
\label{eq:app}
\scalebox{0.85}{
\begin{math}
\begin{aligned}
\nonumber
\mathcal{D}_{k1,k2} = &\\
&\left\{\begin{matrix}
inf,  ~~~~~~if~ \Pi _{k1} \cap \Pi _{k2}\neq \varnothing \\ 
\frac{1}{N_{k1}N_{k2}}
\sum_{i \in \Pi _{k1}}
\sum_{j \in \Pi_{k2}}
\big(1-\frac{f^{k1}_{i}f^{k2}_{j}}{\left \| f^{k1}_{i}\right \| \left \| f^{k2}_{j}\right \|} \big),   otherwise
\end{matrix}\right. 
\end{aligned}
\end{math}}
\end{equation}
where for tracklets $T_{k1}$ and $T_{k2}$, $\mathcal{D}_{k1,k2}$ is their distance; $\Pi _{k1}$ and $\Pi _{k2}$ are their temporal ranges; $f^{k1}_{i}$ and $f^{k2}_{j}$ are their appearance features at frame $i$ and $j$, and $N_{k1}$ and $N_{k2}$ are the number of observations within the tracklets, respectively.

Based on $\mathcal{D}$, we applied hierarchical clustering to cluster short-term tracklets to long-term ones.

\vspace{-3.5mm}
\section{Results}

\subsection{Challenge Results}

We obtained the $2^{nd}$ place in the ECCV 2022 Group Dance Challenge (Table~\ref{table:ranking}), and the result details are summerized in Table~\ref{table:dancetrack_challenge}.
Since we directly applied the YOLOX trained by OC-SORT~\cite{cao2022observation} to perform the detection without any fine tuning, our DetA score (\ie, detection performance) is worse than other teams, which further affects our HOTA, AssA and IDF1 scores. We assume that our tracking performance could be further improved if strong detections are provided.

\begin{table}[t]
   \caption{\textbf{The top-3 teams in the ECCV 2022 Group Dance Challenge.} The data rendered in \textbf{Bold} and \underline{Underlined} indicate the best and
   second best results respectively.}
   \centering
   \setlength{\tabcolsep}{1.8pt}
   \footnotesize
   \begin{tabular}{ llccccc}
   \toprule
   Ranking & Team Name & HOTA$\uparrow$ & DetA$\uparrow$ & AssA$\uparrow$ & MOTA$\uparrow$ & IDF1$\uparrow$\\
   \midrule   
   $1^{st}$ place & mfv & \textbf{73.4} & 	\underline{83.7} &	\textbf{64.4} &	92.1 &	\textbf{76.0} \\
   $2^{nd}$ place & C-BIoU (Ours) & \underline{66.6} & 81.3 & \underline{54.7} & 92.3 &	\underline{71.4}\\ 
   $3^{rd}$ place& ymzis69 & 64.6 &	82.5 &	50.7 &	92.3 &	64.5 \\
   \bottomrule
   \end{tabular}
   \label{table:ranking}
\end{table}

\begin{table}[t]
   \caption{\textbf{Ablation experiments on the DanceTrack validation set~\cite{sun2022dancetrack}}. Where ``C.M.''  and ``Mo.'' represent the cascaded matching and motion estimation, respectively. We remove the cascaded matching and motion estimation in Fig.~\ref{fig:C-BIoU_framework} to construct \colorbox{lavender}{a unified framework} for the IoU, GIoU~\cite{giou2019}, DIoU~\cite{zheng2020distance}, and BIoU. }
   \centering
   \setlength{\tabcolsep}{0.7pt}
   \footnotesize
   \begin{tabular}{ l c c ccccc}
   \toprule
   Tracker  ~~& C.M. ~~~&Mo. ~~ & HOTA$\uparrow$ & DetA$\uparrow$ & AssA$\uparrow$ & MOTA$\uparrow$ & IDF1$\uparrow$\\
   \midrule
   \hline
   \multicolumn{8}{l}{DanceTrack Validation Set~\cite{sun2022dancetrack}. Using Oracle Detections. }   \\
   \rowcolor{lavender}
   IoU Tracker &\xmark & \xmark & 76.6 & 97.5 & 60.2 & 99.2 & 73.6\\
   \rowcolor{lavender}
   GIoU Tracker &\xmark & \xmark &77.1 &\textbf{97.6} & 60.9  & 99.2& 74.0\\
   \rowcolor{lavender}
   DIoU Tracker &\xmark & \xmark & 75.1 & 97.0 & 58.2 & 99.2 & 72.9\\
   \rowcolor{lavender}
   BIoU Tracker &\xmark & \xmark & 80.0 & 97.5 & 65.7 & \textbf{99.3} & 78.2\\
   C-BIoU Tracker &\cmark & \xmark & 80.2 & 97.5 & 65.9 & \textbf{99.3} & 79.3\\
   \textbf{C-BIoU Tracker} &\cmark & \cmark &  \textbf{81.7} & \textbf{97.6} & \textbf{68.4} & \textbf{99.3} & \textbf{80.5}\\
   \hline
   \bottomrule
   \end{tabular}
   \label{table:ablation_on_oracle_detection}
\end{table}

\subsection{Ablation Experiments}
\label{sec:ablation}

\subsubsection{Effect of Each Module in the C-BIoU Tracker}
\label{sec:ablation_module}

\begin{figure}[h!]
   \centering
   \includegraphics[width=0.75\linewidth]{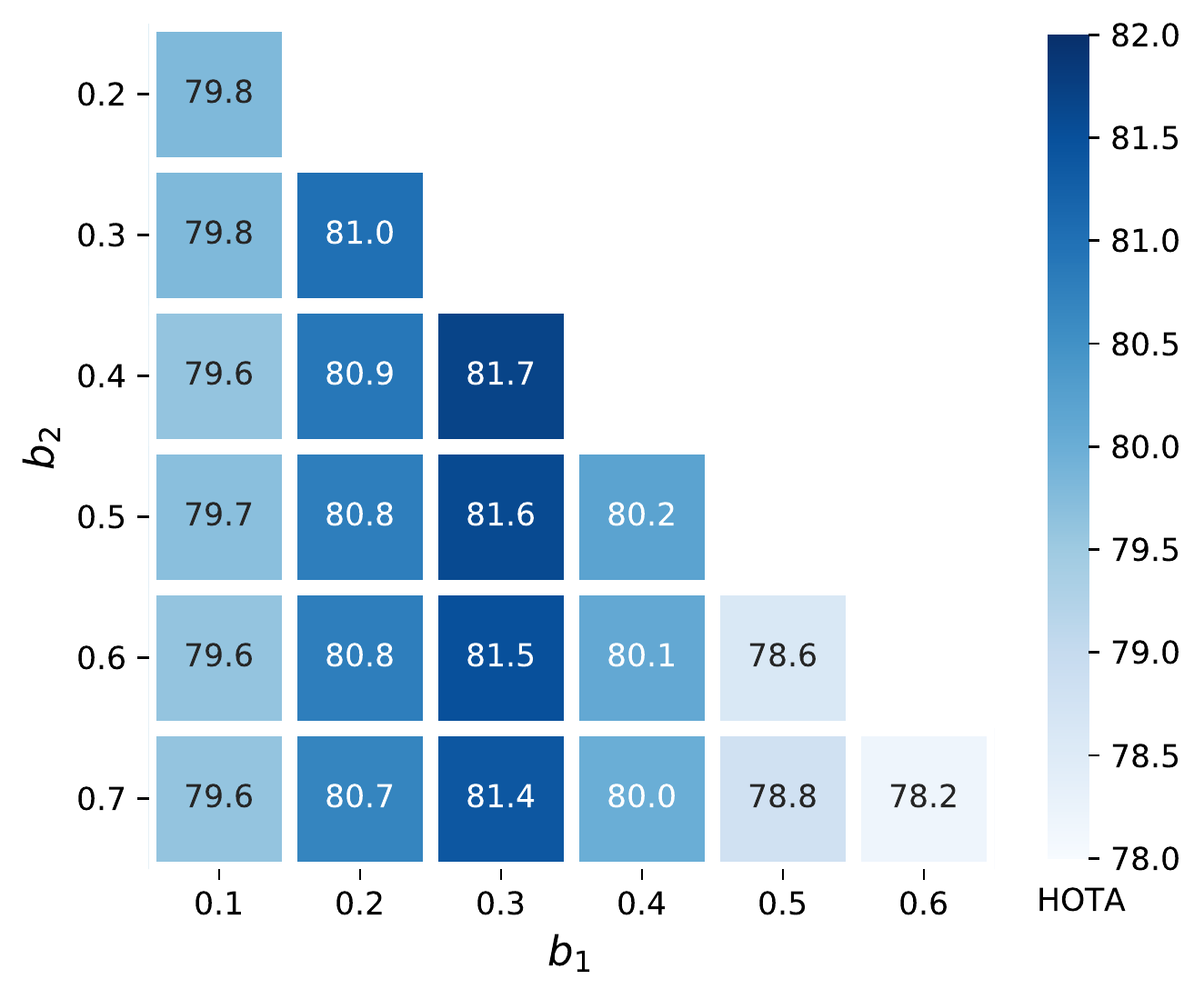}
   \caption{\textbf{Results of applying various buffer-scale combinations on the DanceTrack validation set~\cite{sun2022dancetrack}.} For buffer scales $b_{1}$ and $b_{2}$, since we have $b_{1}<b_{2}$, we only check the lower triangle of the combination matrix.}
   \label{fig:buffer_scales}
\end{figure}

Table~\ref{table:ablation_on_oracle_detection} shows the influence of each module in our C-BIoU tracker. 
 Using the same framework, the tracker equipped with BIoU achieves a higher HOTA score than other trackers equipped with IoU, GIoU~\cite{giou2019}, or DIoU~\cite{zheng2020distance}. Although the GIoU and DIoU can incorporate non-overlapping boxes for geometric consistency measurement, they may not generate comparable results as our BIoU does. Integrating cascaded matching and BIoU can slightly improve the performance as compared to using BIoU alone, with a HOTA gain of $0.2$. According to the results, motion estimation plays an important role in our C-BIoU tracker. Since our BIoU can compensate the matching space for incorrect motion estimation, using a simple motion estimation (\ie, averaging previous motions) yields better HOTA scores than that without using motion estimation.

\subsubsection{Effect of Buffer Scales in the C-BIoU Tracker}

In our C-BIoU tracker, the buffer scales $b_{1}$ and $b_{2}$ are critical hyperparameters. Here, we perform ablation studies to investigate how buffer scales affect the tracking performance. 
On the DanceTrack validation set~\cite{sun2022dancetrack}, we form the combination of $b_{1}$ and $b_{2}$ ranging from $0.1$ to $0.7$ and evaluate their tracking performance. Since we have $b_{1}<b_{2}$, we only need to check $21$ combinations. As shown in Fig.~\ref{fig:buffer_scales}, the combination of $[0.3, 0.4]$ gives the maximum HOTA score. In real practice, we perform a similar approach to select the best combination on the training dataset and apply them to the test dataset.

\section{Conclusion}

We present a novel Cascaded-Buffered IoU (C-BIoU) tracker to track multiple objects that have indistinguishable appearances and irregular motions. The good performance of our C-BIoU tracker can be attributed to its buffered matching space, which mitigates the effect of irregular motions in two aspects: one is to directly match identical but non-overlapping detections and tracks in adjacent frames, and the other is to compensate for the motion estimation bias in the matching space.

{\small
\bibliographystyle{ieee_fullname}
\bibliography{egbib}
}

\end{document}